\documentclass[nohyperref]{article}

\usepackage{amsmath,amsfonts,bm}

\def\eqref#1{equation~\ref{#1}}

\def\1{\bm{1}}

\DeclareMathAlphabet{\mathsfit}{\encodingdefault}{\sfdefault}{m}{sl}
\SetMathAlphabet{\mathsfit}{bold}{\encodingdefault}{\sfdefault}{bx}{n}

\def\gC{{\mathcal{C}}}

\def\gL{{\mathcal{L}}}

\def\gP{{\mathcal{P}}}
\def\gQ{{\mathcal{Q}}}

\def\sR{{\mathbb{R}}}
\def\sS{{\mathbb{S}}}

\newcommand{\E}{\mathbb{E}}

\usepackage{microtype}
\usepackage{graphicx}
\usepackage{subfigure}
\usepackage{caption}
\usepackage{booktabs} 
\usepackage{makecell}
\usepackage{multicol}
\usepackage{multirow}
\usepackage{float}
\usepackage{enumitem}

\usepackage{hyperref}

\DeclareMathOperator{\MMD}{MMD}

\usepackage[accepted]{icml2022}

\usepackage{amsmath}
\usepackage{amssymb}
\usepackage{mathtools}
\usepackage{amsthm}
\usepackage{pifont}
\usepackage[capitalize,noabbrev]{cleveref}

\theoremstyle{plain}

\theoremstyle{definition}

\theoremstyle{remark}

\usepackage[textsize=tiny]{todonotes}

\icmltitlerunning{
How Robust is Your Fairness? Evaluating and Sustaining Fairness under Unseen Distribution Shifts
}

\begin{document}

\twocolumn[
\icmltitle{
How Robust is Your Fairness? \\
Evaluating and Sustaining Fairness under Unseen Distribution Shifts
}



\icmlsetsymbol{equal}{*}

\begin{icmlauthorlist}
\icmlauthor{Haotao Wang}{1}
\icmlauthor{Junyuan Hong}{2}
\icmlauthor{Jiayu Zhou}{2}
\icmlauthor{Zhangyang Wang}{1}
\end{icmlauthorlist}

\icmlaffiliation{1}{University of Texas at Austin}
\icmlaffiliation{2}{Michigan State University}

\icmlcorrespondingauthor{Haotao Wang}{htwang@utexas.edu}

\icmlkeywords{Machine Learning, ICML}

\vskip 0.3in
]



\printAffiliationsAndNotice{}  

\begin{abstract}
Increasing concerns have been raised on deep learning fairness in recent years. Existing fairness-aware machine learning methods mainly focus on the fairness of in-distribution data.  
However, in real-world applications, it is common to have distribution shift between the training and test data.
In this paper, we first show that the fairness achieved by existing methods can be easily broken by slight distribution shifts. 
To solve this problem, we propose a novel fairness learning method termed CUrvature MAtching (CUMA), which can achieve robust fairness generalizable to unseen domains with unknown distributional shifts. 
Specifically, CUMA enforces the model to have similar generalization ability on the majority and minority groups, by matching the loss curvature distributions of the two groups.
We evaluate our method on three popular fairness datasets.
Compared with existing methods, CUMA achieves superior fairness under unseen distribution shifts, without sacrificing either the overall accuracy or the in-distribution fairness.
\end{abstract}

\section{Introduction}
\label{sec:intro}

\begin{figure*}[t] 
	\centering
	\setlength{\tabcolsep}{1pt}
	\begin{tabular}{ccc}
        \includegraphics[width=0.33\linewidth]{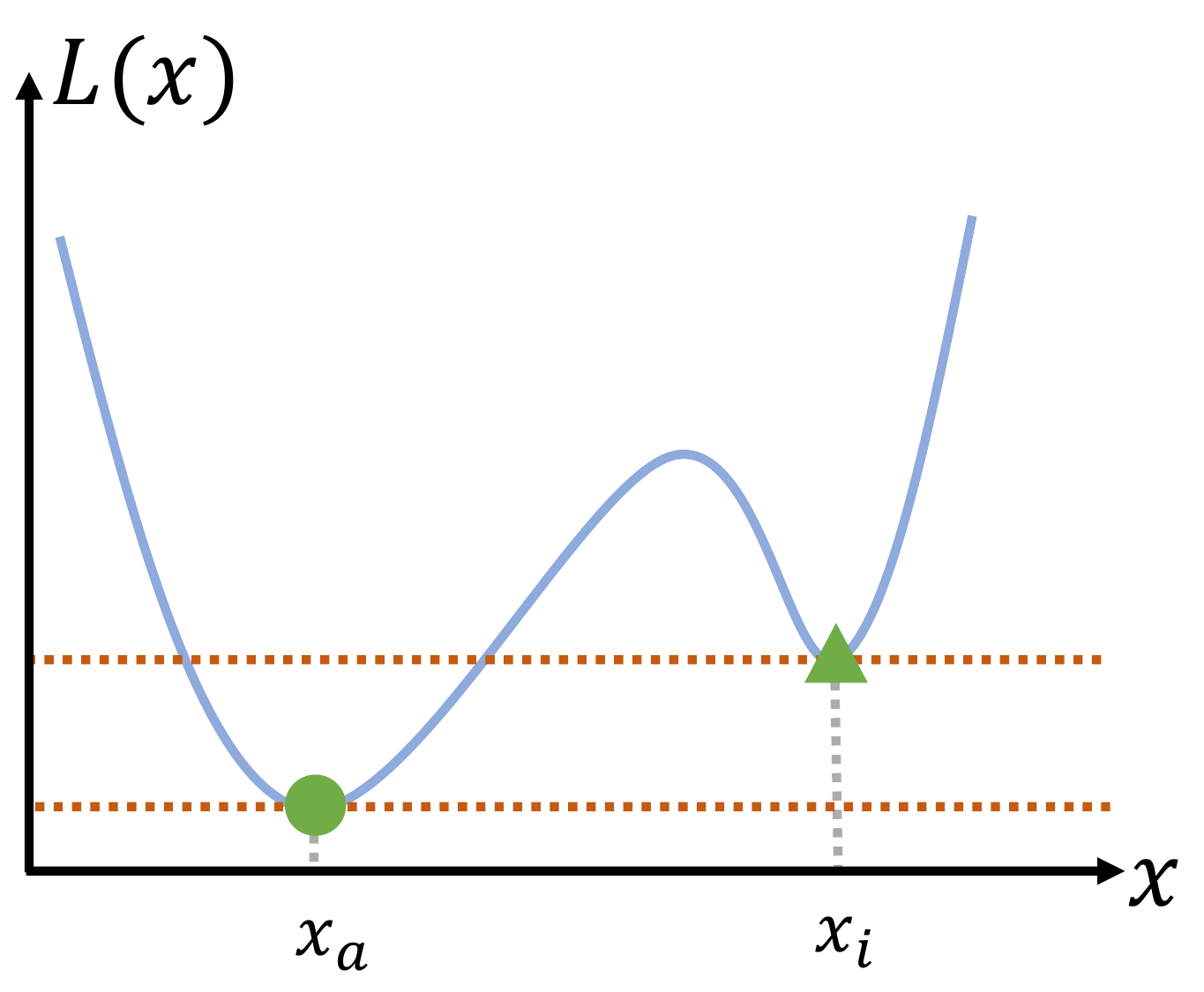} &
		\includegraphics[width=0.33\linewidth]{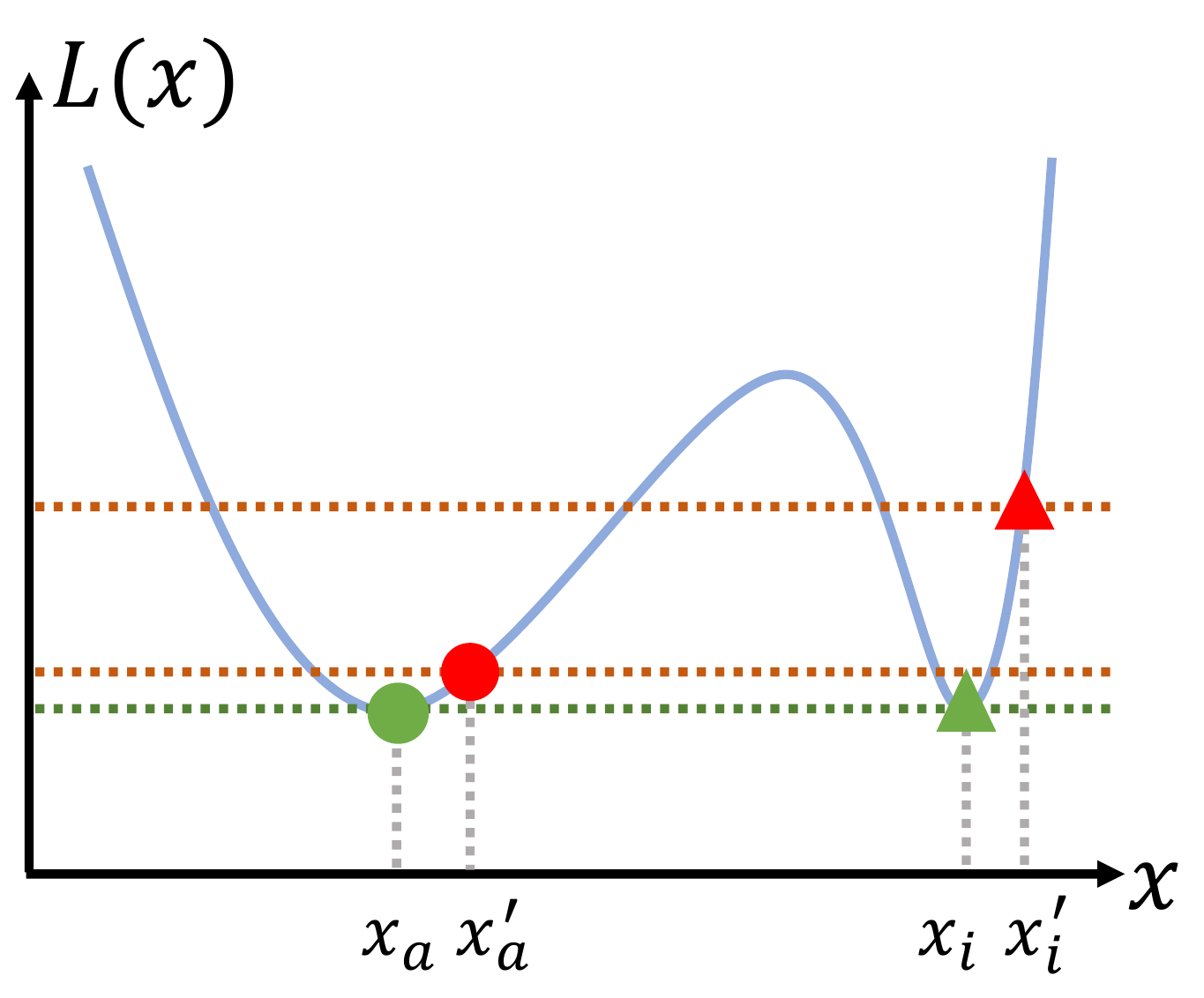} &
		\includegraphics[width=0.33\linewidth]{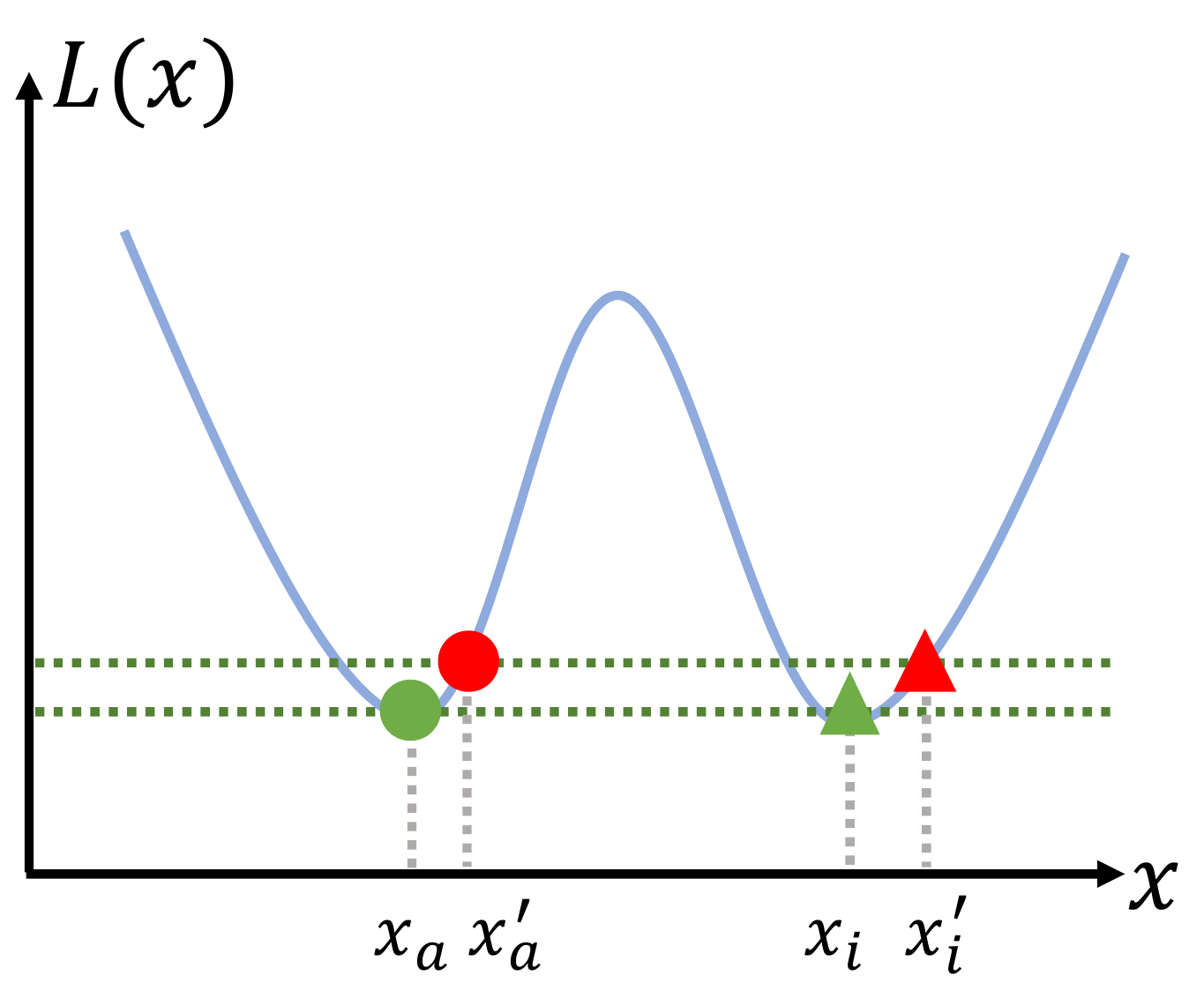} \\
		\makecell{(a) Normal learning\\(Unfair)} &  \makecell{(b) Traditional fairness learning\\(In-distribution fairness)} &  \makecell{(c) Our robust fairness learning\\(In-distribution \& robust fairness)}
	\end{tabular}
	\vspace{-1em}
	\caption{
 Illustrating the achieved fairness of normal training, traditional fair training and our proposed robust fair training algorithms. Horizontal and vertical axes represent input $x$ and corresponding loss value $\gL(x)$, respectively. Solid blue curves show the loss landscapes.
	Circles denote majority data points ($x_a$ and $x_a^{\prime}$), while triangles denote minority data points ($x_i$ and $x_i^{\prime}$). 
	Green points ($x_a$ and $x_i$) are in-distribution data while red ones ($x_a^{\prime}$ and $x_i^{\prime}$) are sampled from test sets with distribution shifts. 
	(a) Normal training results in unfair models: minority group has worse performance (i.e., larger loss values).
	(b) Traditional fairness learning algorithms can achieve in-distribution fairness but not in a robust way: a small distribution shift can break the fairness due to loss curvature biases across different groups. In fact, such learned fair models can have almost the same large bias as the normally trained models when facing distribution shifts. 
	(c) Our robust fairness learning algorithm can simultaneously achieve fairness both on in-distribution data and at distribution shifts, by matching both loss values and loss curvatures across different groups.
	}
	\vspace{-0.5em}
	\label{fig:teaser}
\end{figure*}

With the wide deployment of deep learning in modern business applications concerning individual lives and privacy, there naturally emerge concerns on machine learning fairness \citep{housebig,executive2016big,smuha2019eu}. 
Although research efforts on various fairness-aware learning algorithms have been carried out \citep{edwards2015censoring, hardt2016equality,du2020fairness},
most of them focus only on equalizing model performance across different groups on \textit{in-distribution} data. 

Unfortunately, in real-world applications, one commonly encounter data with unforeseeable distribution shifts from model training. 
It has been shown that deep learning models have drastically degraded performance \citep{hendrycks2019robustness,hendrycks2020many, hendrycks2021nae, taori2020measuring} and show unpredictable behaviors \citep{qiu2019adversarial,yan2021cifs} under unseen distribution shifts. 
Intuitively speaking, previous fairness learning algorithms aim to optimize the model to a local minimum where data from majority and minority groups have similar average loss values (and thus similar in-distribution performance). 
However, those algorithms do not take into consideration the stability or ``robustness" of their found fairness-aware minima. Taking object detection in self-driving cars for example, it might have been calibrated over high-quality clear images to be ``fair" with different pedestrian skin colors; however such fairness may break down when applied to data collected in adverse visual conditions, such as inclement weather, poor lighting, or other digital artifacts.
Our experiments also find that previous state-of-the-art fairness algorithms would be jeopardized if distributional shifts are present in test data.
The above findings beg the following question: 
\begin{center}
\textit{
Using the currently available training set, how to achieve robust fairness that is generalizable to unseen domains with unpredictable distribution shifts?
}
\end{center}


To solve this problem, we decompose it into the following two-step objectives and achieve them one by one:
\begin{enumerate}[leftmargin=*]
\item The minority and majority groups should have similar prediction \textbf{accuracy} on \textit{in-distribution} data. This is usually attained by traditional in-distribution fairness goals.
\item The minority and majority groups should have similar \textbf{robustness} under \textit{unseen distributional shifts}. In this context, the ``robustness" refers to model performance gap between in-distribution and unseen out-of-distribution data: the larger gap the weaker.
\end{enumerate}
The first objective is well studied and can be achieved by existing fairness learning methods such as adversarial training \cite{edwards2015censoring, hardt2016equality,du2020fairness}.
In this paper, we focus our efforts on addressing the second objective, which has been much less studied. We present empirical evidence that the fairness achieved by existing in-distribution oriented methods can be easily compromised by even slight distribution shifts.

Next, to mitigate this fragility, we note that model robustness against distributional shift is often found to be highly correlated with the loss curvature \citep{bartlett2017spectrally,weng2018evaluating}.
Our experiments further observed that, the local loss curvature of minority group is often much larger than that of majority group, which explains their discrepancy of robustness. 
Motivated by the above, we propose a new fairness learning algorithm, termed \textbf{Curvature Matching (CUMA)}, to robustify the fairness. 
Specifically, CUMA enforces the model to have similar robustness on the majority and minority groups, by matching the loss curvature distributions of the two groups.
As a plug-and-play modular, CUMA can be flexibly combined with existing in-distribution fairness learning methods, such as adversarial training, to fulfil our overall goal of robust fairness.
We illustrate the core idea of CUMA and its difference compared with traditional in-distribution fairness methods in Figure \ref{fig:teaser}.

We evaluate our method on three popular fairness datasets: Adult, C\&C, and CelebA.
Experimental results show that CUMA achieves significantly more robust fairness against unseen  distribution  shifts, without sacrificing either overall accuracy or the in-distribution fairness, compared to traditional fairness learning methods.



\section{Preliminaries}

\subsection{Machine Learning Fairness} \label{sec:pre}

\paragraph{Problem Setting and Metrics}
Machine learning fairness can be generally categorized into individual fairness and group fairness \citep{du2020fairness}. Individual fairness requires similar inputs to have similar predictions \citep{dwork2012fairness}. Compared with individual fairness, group fairness is a more popular setting and thus the focus of our paper.
Given input data $X\in\sR^n$ with sensitive attributes $A\in\{0,1\}$ and their corresponding ground truth labels $Y\in\{0,1\}$, group fairness requires a learned binary classifier $f(\cdot;\theta):\sR^n \rightarrow \{0,1\}$ parameterized by $\theta$ to give equally accurate predictions (denoted as $\hat{Y} \coloneqq f(X)$) on the two groups with $A=0$ and $A=1$. 
Multiple fairness criteria have been defined in this context. 
Demographic parity (DP) \citep{edwards2015censoring} requires identical ratio of positive predictions between two groups: $P(\hat{Y}=1|A=0) = P(\hat{Y}=1|A=1)$.
Equalized Odds (EO) \citep{hardt2016equality} requires identical false positive rates (FPRs) and false negative rates (FNRs) between the two groups: $P(\hat{Y} \neq Y|A=0, Y=y) = P(\hat{Y} \neq Y|A=1, Y=y), \forall y \in \{0,1\}$. 
Based on these fairness criteria, quantified metrics are defined to measure fairness. 
For example, the EO distances \citep{madras2018learning} is defined as follows:
\begin{align}
\Delta_{EO} := \sum_{y\in\{0,1\}}|&P(\hat{Y} \neq Y|A=0, Y=y) - \nonumber  \\
&P(\hat{Y} \neq Y|A=1, Y=y)| 
\label{eq:eo}
\end{align}


\paragraph{Bias Mitigation Methods}
Many methods have been proposed to mitigate model bias. Data pre-processing methods such as re-weighting \citep{kamiran2012data} and data-transformation \citep{calmon2017optimized} have been used to reduce discrimination before model training. In contrast, \cite{hardt2016equality} and \cite{zhao2017men} propose post-processing methods to calibrate model predictions towards a desired fair distribution after model training.
Instead of pre- or post-processing, researchers have explored to enhance fairness during training.
For example, \cite{madras2018learning} uses a adversarial training technique and shows the learned fair representations can transfer to unseen target tasks.
The key technique, adversarial training \citep{edwards2015censoring}, was designed for feature disentanglement on hidden representations such that sensitive \citep{edwards2015censoring} or domain-specific information \citep{ganin2016domain} will be removed while keeping other useful information for the target task.
The hidden representations are typically the output of intermediate layers of neural networks \citep{ganin2016domain,edwards2015censoring,madras2018learning}.
Instead, methods, like adversarial debiasing \citep{zhang2018mitigating} and its simplified version \citep{wadsworth2018achieving}, directly apply the adversary on the output layer of the classifier, which also promotes the model fairness.
Observing the unfairness due to ignoring the worst learning risk of specific samples, \citet{hashimoto2018fairness} proposes to use distributionally robust optimization which provably bounds the worst-case risk over groups.
\citet{creager2019flexibly} proposes a flexible fair representation learning framework based on VAE \citep{kingma2013auto}, that can be easily adapted for different sensitive attribute settings during run-time.
\citet{sarhan2020fairness} uses orthogonality constraints as a proxy for independence to disentangles the utility and sensitive representations.
\citet{martinez2020minimax} formulates group fairness with multiple sensitive attributes as a multi-objective learning problem and proposes a simple optimization algorithm to find the Pareto optimality. 
Another line of research focuses on learning unbiased representations from biased ones \citep{bahng2020learning, nam2020learning}. 
\citet{bahng2020learning} proposes a novel framework to learn unbiased representations by explicitly enforcing them to be different from a set of pre-defined biased representations.
\citet{nam2020learning} observes that data bias can be either benign or malicious, and removing malicious bias along can achieve fairness.
\citet{li2019repair} jointly learns a data re-sampling weight distribution that penalizes easy samples and network parameters.
\citet{li2019fair} scaled by higher-order power to re-emphasize the loss of minority samples (or nodes) in distributed learning.
\citet{agarwal2018reductions} formulates a fairness-constrained optimization to train a randomized classifier which is provably accurate and fair.
\citet{quadrianto2019discovering} casts the sensitive information removal problem as a data-to-data translation problem with unknown target domain.

\paragraph{Applications in Computer Vision}
When many fairness metrics and debiasing algorithms are designed for general learning problems as aforementioned, there are a line of research and applications focusing on fairness-encouraged computer vision tasks.
For instance, 
\citet{buolamwini2018gender} shows current commercial gender-recognition systems have substantial accuracy disparities among groups with different genders and skin colors.
\citet{wilson2019predictive} observe that state-of-the-art segmentation models achieve better performance on pedestrians with lighter skin colors. 
In \citep{shankar2017no,de2019does}, it is found that the common geographical bias in public image databases can lead to strong performance disparities among images from locales with different income levels.
\citet{nagpal2019deep} reveal that the focus region of face-classification models depends on people's ages or races, which may explain the source of age- and race-biases of classifiers.
On the awareness of the unfairness, many efforts have been devoted to mitigate such biases in computer vision tasks.
\citet{wang2019balanced} shows the effectiveness of adversarial debiasing technique \citep{zhang2018mitigating} in fair image classification and activity recognition tasks.
Beyond the supervised learning, FairFaceGAN \citep{hwang2020fairfacegan} is proposed to prevent undesired sensitive feature translation during image editing. 
Similar ideas have also been successfully applied to visual question answering \citep{park2020fair}. 

\paragraph{Fairness under distributional shift}
Recently, several papers have investigated the fairness learning problem under distributional shift \cite{mandal2020ensuring,zhang2021farf,rezaei2021robust,singh2021fairness,dailabel}. Although these works are relevant with ours, there are significant differences in the problem settings. 
\citet{zhang2021farf} studied the problem of enforcing fairness in online learning, where the training distribution constantly shifts. 
The authors proposed to adapt the model to be fair on the current \textit{known} data distribution. In contrast, our work aims to generalize fairness learned on current distribution to \textit{unknown} and \textit{unseen} target distributions. In our setting, the algorithm can not access any training data from the unknown target distributions.
\citet{rezaei2021robust} studied to preserve fairness under covariate shift. However, their method requires unlabeled data from the target distribution. In other words, they assume the target distribution to be \textit{known}. In contrast, our method is more general and works on \textit{unknown} target distributions.
\citet{singh2021fairness} also studied to preserve fairness under covariate shift. However, their method is based on model adaptation and requires the existence of a joint causal graph to represent the data distribution for all domains. Our method, however, does not require such requirement and generally works on any unseen target distributions. 
\citet{dailabel} studies fairness under label distributional shift, while we focus on covariate shift.

\subsection{Model Robustness and Smoothness}
Model generalization ability and robustness has been shown to be highly correlated with model smoothness \citep{moosavi2019robustness,weng2018evaluating}.
\citet{weng2018evaluating} and \citet{guo2018sparse} use local Lipschitz constant to estimate model robustness against small perturbations on inputs within a hyper-ball. \citet{moosavi2019robustness} proposes to improve model robustness by adding a curvature constraint to encourage model smoothness.
\citet{miyato2018virtual} approximates model local smoothness by the spectral norm of Hessian matrix, and improves model robustness against adversarial attacks by regularizing model smoothness.

\section{The Challenge of Robust Fairness}
\label{sec:challenge}

In this section, we show that the current state-of-the-art in-distribution fairness learning methods suffer significant performance drop under unseen distribution shifts.
Specifically, we train the model using normal training (denoted as ``Normal" in Table \ref{tab:teaser}), AdvDebias \citep{zhang2018mitigating} and LAFTR \cite{madras2018learning} on Adult \cite{kohavi1996scaling} dataset (i.e., US Census data before 1996). 
We evaluate the $\Delta_{EO}$ on the original Adult test set and the 2015 subset of Folktables datase \cite{ding2021retiring} (i.e., US Census data in 2015) respectively, in order to check whether the fairness achieved on in-distribution data is preserved under the temporal distribution shift.
The results are shown in Table \ref{tab:teaser}.

As we can see, LAFTR and AdvDebias successfully improve the in-distribution fairness compared with normal training. 
However, both methods suffer significant performance drop in terms of $\Delta_{EO}$ under the temporal distribution shift. 
Moreover, under the distribution shift, the $\Delta_{EO}$ achieved by LAFTR and AdvDebias are almost the same with that of normal training. In other words, the models trained by LAFTR and AdvDebias are almost as unfair as a normally trained model under this naturally occurring distribution shift.

\begin{table}[ht]
\begin{center}
\caption{
Existing in-distribution fairness learning methods suffer significant performance drop under distribution shifts.
All methods are trained on Adult dataset (i.e., US Census data before 1996) with ``Sex'' as the sensitive attribute. 
The best and second-best metrics are shown in bold and underlined, respectively. Mean and standard deviation over three random runs are shown for our method.}
\vspace{-0.5em}
\resizebox{\linewidth}{!}{
\begin{tabular}{|c|c|c|}
\hline
\multirow{3}{*}{Method} & In-distribution fairness & \makecell{Robust fairness under\\unseen distribution shift} \\
\cline{2-3}
& \makecell{$\Delta_{EO}$ ($\downarrow$) on US Census data\\before 1996\\(i.e., the in-distribution test set)} & \makecell{$\Delta_{EO}$ ($\downarrow$) on US Census data\\in 2015} \\
\hline\hline
Normal &  15.45 &  14.65 \\
LAFTR &  11.96 &  14.80 \\
AdvDebias & \underline{5.92} &  \underline{12.35} \\
\hline
Ours & \textbf{4.77}$\pm0.34$ & \textbf{8.20}$\pm$1.26 \\
\hline
\end{tabular}
}
\label{tab:teaser}
\vspace{-1em}
\end{center}
\end{table}

\section{Curvature Matching: Towards Robust Fairness under Unseen Distributional Shifts}

In this section, we present our proposed solution for the robust fairness challenge described in Section \ref{sec:challenge}. 

\subsection{Loss Curvature as the Measure for Robustness}

Before introducing our robust fairness learning method, we need to first define the measure for model robustness under unseen distributional shifts. 

Consider a binary classifier $f(\cdot;\theta)$ trained on two groups of data $X_1$ and $X_2$. Our goal is to define a metric to measure the gap of model robustness between the two groups.
Previous research \citep{guo2018sparse,weng2018evaluating} has shown both theoretically and empirically that deep model robustness scales with its model smoothness.
Motivated by the above, we use the spectral norm of Hessian matrix to approximate local smoothness as a measure of model robustness. Specifically, given an input $x$, the Hessian matrix $H(x)$ is defined as the second-order gradient of $\mathcal{L}(x)$ with respect to model weights $\theta$: $H(x) = \nabla^2_{\theta} \mathcal{L}(x)$.
The approximated local curvature $\gC(x)$ at point $x$ is thus defined as:
\begin{equation} \label{eq:curvature}
    \gC(x) = \sigma(H(x)),
\end{equation}
where $\sigma(H)$ is the spectral norm (SN) of $H$: $\sigma(H) = \sup_{v:\|v\|_2=1} \|Hv\|_2$.
Intuitively, $\gC(x)$ measures the maximal directional curvature or change rate of the loss function at $x$. Thus, smaller $\gC(x)$ indicates better local smoothness around $x$.

\paragraph{Practical Curvature Approximation}
It is inefficient to directly optimize the loss curvature through Eq. (\ref{eq:curvature}), since it involves high order gradients.\footnote{The Hessian matrix itself involves second order gradients, and backpropagation through Eq. (\ref{eq:curvature}) requires even higher order gradient on top of the Hessian matrix SN.}
To solve this problem, we use a one-shot power iteration method (PIM) for practical approximation of $\gC(x)$ during training.
First we rewrite $\gC(x)$ with the following form: $\gC(x) = \sigma(H(x)) = \|H(x)v\|$, 
where $v$ is the dominant eigenvector with the maximal eigenvalue, which can be calculated by power iteration method. 
In practice, we estimate the dominant eigenvector $v$ by the gradient direction: $\tilde{v} \coloneqq \frac{\text{sign}(g)}{\|\text{sign}(g)\|} \approx v$, where $g=\nabla_{\theta} \gL(x)$.
This is because previous works have observed a large similarity between the dominant eigenvector and the gradient direction \citep{miyato2018virtual,moosavi2019robustness}. We further approximate Hessian matrix by finite differentiation on gradients: $H(x)v \approx \frac{\nabla_\theta \gL(x+hv) - \nabla_\theta \gL(x)}{h}$ where $h$ is a small constant. As a result, the final approximation of curvature smoothness is 
\begin{equation}
\label{eq:curvature_aprx}
\begin{split}
    \tilde{\gC}(x) \coloneqq \frac{\|\nabla_\theta \gL(x+h\tilde{v}) - \nabla_\theta \gL(x)\|}{|h|} \approx \gC(x).
\end{split}
\end{equation}

\begin{table*}[ht]
\begin{center}
\caption{Results on Adult dataset with ``Sex'' as the sensitive attribute. The best and second-best metrics are shown in bold and underlined, respectively. Mean and standard deviation over three random runs are shown for CUMA.}
\vspace{-0.5em}
\resizebox{\linewidth}{!}{
\begin{tabular}{|c|c|c|c|c|}
\hline
\multirow{3}{*}{Method} & \multicolumn{2}{c|}{\makecell{The original Adult test set\\(US Census data before 1996)}} & \multicolumn{1}{c|}{\makecell{Folktables 2014 subset\\(US Census data in 2014)}} & \multicolumn{1}{c|}{\makecell{Folktables 2015 subset\\(US Census data in 2015)}} \\
\cline{2-5}
& \multirow{2}{*}{Accuracy ($\uparrow$)} &  $\Delta_{EO}$ ($\downarrow$) & $\Delta_{EO}$ ($\downarrow$) & $\Delta_{EO}$ ($\downarrow$) \\
\cline{3-5}
& & \multicolumn{1}{c|}{\textit{In-distribution fairness}} & \multicolumn{2}{c|}{\textit{Robust fairness under distribution shifts}}  \\
\hline\hline
Normal & \textbf{86.11} & 15.45 & 14.28 & 14.65 \\
AdvDebias & {85.17} & \underline{5.92} & \underline{8.25} & \underline{10.16} \\
LAFTR & \underline{85.97} & 11.96 & 13.95 & 14.80 \\
CUMA & 85.30$\pm0.73$  & \textbf{4.77}$\pm0.34$ & \textbf{6.33}$\pm$0.94 & \textbf{8.20}$\pm$1.26 \\
\hline
\end{tabular}
}
\label{tab:adult}
\end{center}
\end{table*}

\subsection{Curvature Matching}
\label{sec:cuma}

Equipped with the practical curvature approximation, now we can match the curvature distribution of the two groups by minimizing their maximum-mean-discrepancy (MMD) \cite{gretton2012kernel} distance. 
Suppose $\tilde{\gC}(X_1) \sim \gQ_1$ and $\tilde{\gC}(X_2) \sim \gQ_2$, we define the curvature matching loss functions as:
\begin{equation}
\label{eq:Lcm}
\begin{split}
\mathcal{L}_{cm} = \MMD^2(\gQ_1, \gQ_2),
\end{split}
\end{equation}
The MMD distance, which is widely used to measure the distance between two high-dimensional distributions in deep learning \citep{li15gmmn,li2017mmd,binkowski2018demystifying}, is defined as 
\begin{align}
\MMD^2(\gP,\gQ) = &\E_\gP[k(X,X)] - \nonumber \\
&2\E_{\gP,\gQ}[k(X,Y)] + \E_\gQ[k(Y,Y)]
\end{align}
where $X \sim \gP$, $Y \sim \gQ$ and $k(\cdot,\cdot)$ is the kernel function. 
In practice, we use finite samples from $\gP$ and $\gQ$ to statistically estimate their MMD distance:
\begin{align}
    &\MMD^2(\gP, \gQ) = \frac{1}{M^2} \sum_{i=1}^{M}\sum_{i'=1}^{M} k(x_i, x_{i'}) \nonumber \\
    &-\frac{2}{MN} \sum_{i=1}^{M}\sum_{j=1}^{N} k(x_i, y_j) +\frac{1}{N^2} \sum_{j=1}^{N}\sum_{j'=1}^{N} k(y_j, y_{j'})
\end{align}
where $\{x_i\sim \gP\}_{i=1}^M$, $\{y_j \sim \gQ \}_{j=1}^N$, and we use the mixed RBF kernel function $k(x,y)=\sum_{\sigma \in \sS}e^{-\frac{\|x-y\|^2}{2\sigma^2}}$ with hyperparameter $\sS=\{1,2,4,8,16\}$. 

As a side note, MMD has been previously used in fairness learning. \citet{quadrianto2017recycling} defines a more general fairness metric using MMD distance, and shows DP and EO to be the spatial cases of their unified metric. Their paper, however, still focuses on the in-distribution fairness.
In contrast, our CUMA minimizes the MMD distance on the curvature distributions to achieve robust fairness.

\begin{figure}[ht]
\includegraphics[width=\columnwidth]{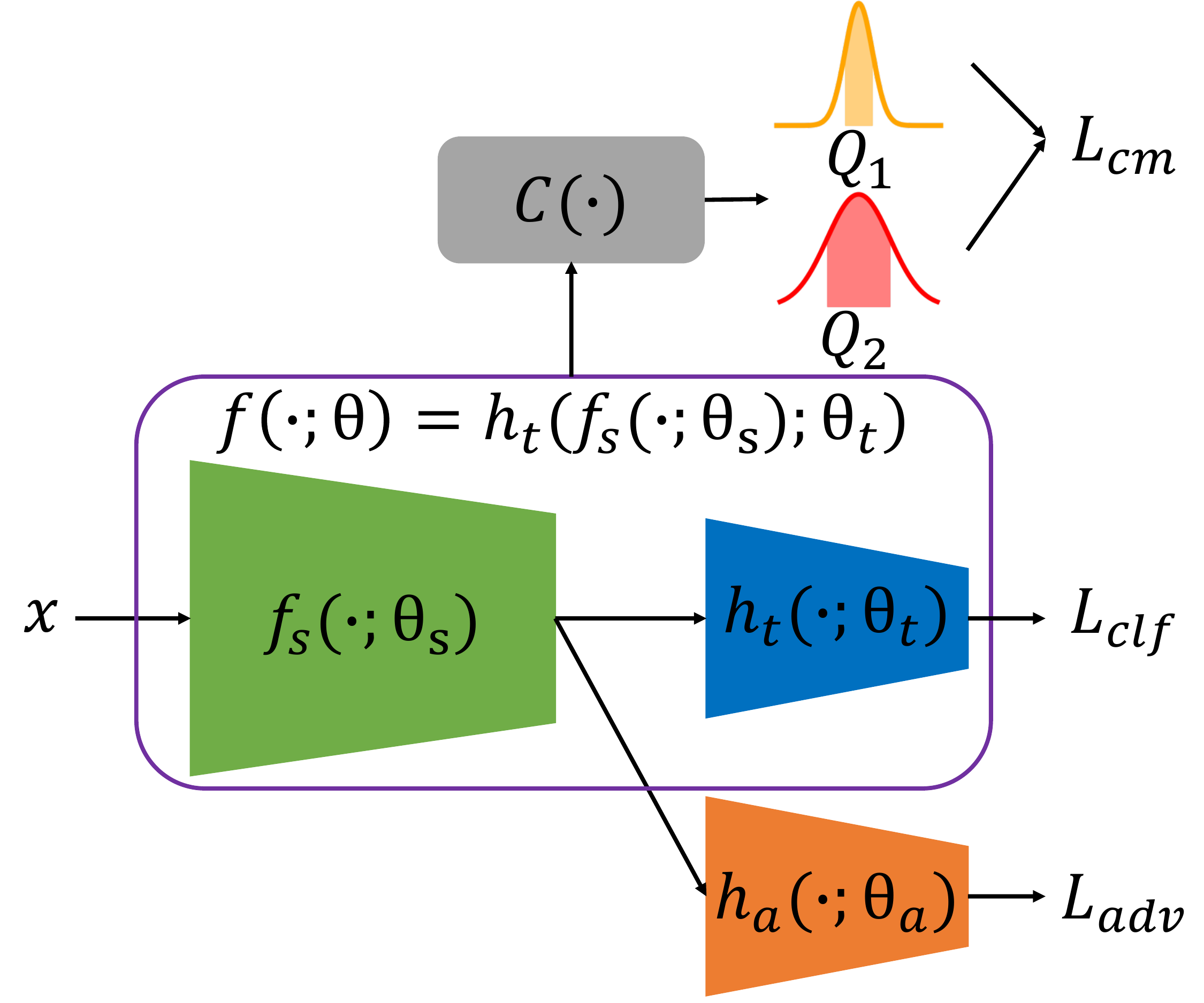}
\caption{
The overall framework of CUMA. $x$ is the input sample. $h_t$ is the utility head for the target task. $h_a$ is the adversarial head to predict sensitive attributes. $f_s$ is the shared backbone.
$\gC(\cdot)$ is the curvature estimation function, as defined in Eq.~(\ref{eq:curvature}).
$\gQ_1$ and $\gQ_2$ are local curvature distributions of majority and minority groups, respectively.
$\mathcal{L}_{cm}$, $\mathcal{L}_{clf}$ and $\mathcal{L}_{adv}$ are three loss terms as defined in Eq.~(\ref{eq:Lcm}), (\ref{eq:Lclf}), and (\ref{eq:Ladv}), respectively.
} 
\label{fig:framework}
\end{figure}

Back to our method. After defining $\mathcal{L}_{cm}$, we add it to the traditional adversarially fair training \citep{ganin2016domain, madras2018learning} loss function as a regularizer, in order to attain both in-distribution fairness and robust fairness. 
As illustrated in Figure~\ref{fig:framework}, our model follows the same ``two-head'' structure as traditional adversarial learning frameworks \citep{ganin2016domain, madras2018learning}, where $h_t$ is the utility head for the target task, $h_a$ is the adversarial head to predict sensitive attributes, and $f_s$ is the shared backbone.\footnote{Thus the binary classifier $f(\cdot;\theta)=h_t(f_s(\cdot; \theta_s); \theta_t)$, with $\theta=\theta_t \cup \theta_s$.}
Suppose for each sample $x_i$, the sensitive attribute is $a_i$ and the corresponding target label is $y_i$, then our overall optimization problem can be written as:
\begin{equation}
\label{eq:overall_loss}
\begin{split}
    \min_{\theta_s,\theta_t}\max_{\theta_a}\mathcal{L} = \min_{\theta_s,\theta_t}\max_{\theta_a}(\mathcal{L}_{clf} - \alpha \mathcal{L}_{adv} + \gamma \mathcal{L}_{cm})
\end{split}
\end{equation}
where 
\begin{align}
&\mathcal{L}_{clf}=\frac{1}{N}\sum_{i=1}^{N} \ell(h_t(f_s(x_i; \theta_s); \theta_t),y_i), \label{eq:Lclf} \\ &\mathcal{L}_{adv}=\frac{1}{N}\sum_{i=1}^{N} \ell(h_a(f_s(x_i;\theta_s);\theta_a),a_i), \label{eq:Ladv}
\end{align}
$\ell(\cdot,\cdot)$ is the cross-entropy loss function, $\alpha$ and $\gamma$ are trade-off hyperparameters, and $N$ is the number of training samples.

\section{Experiments}

\begin{table*}[ht]
\begin{center}
\caption{Results on CelebA dataset with ``Chubby'' as the sensitive attribute. The best and second-best metrics are shown in bold and underlined, respectively.}
\vspace{-0.5em}
\resizebox{\linewidth}{!}{
\begin{tabular}{|c|c|c|c|c|}
\hline
\multirow{3}{*}{Method} & \multicolumn{2}{c|}{\makecell{In-distribution test set\\(Young and high visual quality face images)}} & \multicolumn{1}{c|}{\makecell{Old face images\\with high visual quality}} & \multicolumn{1}{c|}{\makecell{Young face images\\under sever JPEG compression}} \\
\cline{2-5}
& \multirow{2}{*}{Accuracy ($\uparrow$)} &  $\Delta_{EO}$ ($\downarrow$) & $\Delta_{EO}$ ($\downarrow$) & $\Delta_{EO}$ ($\downarrow$) \\
\cline{3-5}
& & \multicolumn{1}{c|}{\textit{In-distribution fairness}} & \multicolumn{2}{c|}{\textit{Robust fairness under distribution shifts}}  \\
\hline\hline
Normal & \textbf{85.76} & 37.42 & 40.52 & 43.01 \\
AdvDebias & \underline{80.31} & 33.25 & 36.54 & \underline{35.86} \\
LAFTR & 79.56 & \textbf{31.02} & \underline{35.88} & 37.46 \\
CUMA & 80.26 & \underline{31.52} & \textbf{33.26} & \textbf{33.12} \\
\hline
\end{tabular}
}
\label{tab:celeba_chubby}
\end{center}
\end{table*}

\begin{table*}[ht]
\begin{center}
\caption{Results on CelebA dataset with ``Eyeglasses'' as the sensitive attribute. The best and second-best metrics are shown in bold and underlined, respectively.}
\vspace{-0.5em}
\resizebox{\linewidth}{!}{
\begin{tabular}{|c|c|c|c|c|}
\hline
\multirow{3}{*}{Method} & \multicolumn{2}{c|}{\makecell{In-distribution test set\\(Young and high visual quality face images)}} & \multicolumn{1}{c|}{\makecell{Old face images\\with high visual quality}} & \multicolumn{1}{c|}{\makecell{Young face images\\under sever JPEG compression}} \\
\cline{2-5}
& \multirow{2}{*}{Accuracy ($\uparrow$)} &  $\Delta_{EO}$ ($\downarrow$) & $\Delta_{EO}$ ($\downarrow$) & $\Delta_{EO}$ ($\downarrow$) \\
\cline{3-5}
& & \multicolumn{1}{c|}{\textit{In-distribution fairness}} & \multicolumn{2}{c|}{\textit{Robust fairness under distribution shifts}}  \\
\hline\hline
Normal & \textbf{85.76} & 43.56 & 42.03 & 40.15 \\
AdvDebias & 78.56 & 33.25 & 37.41 & \underline{35.22} \\
LAFTR & \underline{82.30} & \underline{32.06} & \underline{35.61} & 35.74 \\
CUMA & 81.06 & \textbf{31.15} & \textbf{33.06} & \textbf{32.52} \\
\hline
\end{tabular}
}
\label{tab:celeba_eyeglasses}
\end{center}
\end{table*}

\subsection{Experimental Setup} \label{sec:settings}
\paragraph{Datasets and pre-processing}
Experiments are conducted on three datasets widely used to evaluate machine learning fairness: 
Adult \citep{kohavi1996scaling}, and CelebA \citep{liu2015faceattributes}, and Communities and Crime (C\&C) \citep{redmond2002data}.\footnote{Traditional image classification datasets (e.g., ImageNet) are not directly applicable since they lack fairness attribute labels.}
\underline{Adult} dataset has 48,842 samples with basic personal information such as education and occupation, where 30,000 are used for training and the rest for evaluation. The target task is to predict the person's annual income, and we use ``gender'' (male or female) as the sensitive attribute. The features in Adult dataset are of either continuous (e.g., age) or categorical (e.g. sex) values. We use one-hot encoding on the categorical features and then concatenate them with the continuous ones. We use data whitening on the concatenated features.
\underline{CelebA} has over 200,000 images of celebrity faces, with 40 attribute annotations. The target task is to predict gender (male or female) and the sensitive attributes to protect are ``chubby'' and ``eyeglasses''. We randomly select $10,000$ as training samples and $1,000$ as testing samples. All images are center-cropped and resized to $64\times64$, and pixel values are scaled to $[0,1]$. 
\underline{C\&C} dataset has 1,994 samples with neighborhood population statistics, where 1,500 are used for training and the rest for evaluation. The target task is to predict violent crime per capita, and we use ``RacePctBlack'' (percentage of black population in the neighborhood) and ``FemalePctDiv'' (divorce ratio of female in the neighborhood) as sensitive attributes. All features in C\&C dataset are of continous values in $[0,1]$. To fit in the fairness problem setting, we binarilize the target and sensitive attributes with the top-$30\%$ largest value as the threshold (As a result $\text{P}[A=0]=30\%$ and $\text{P}[Y=0]=30\%$).
We also do data-whitening on C\&C.

\vspace{-1em}
\paragraph{Models}
For C\&C and Adult datasets, we use two-layer MLPs for $f_s$, $h_t$ and $h_a$.
Specifically, suppose the input feature dimension is $d$, then the dimensions of hidden layers in $f_s$ and $h_t$ are $d \rightarrow 100 \rightarrow 64$ and $64 \rightarrow 32 \rightarrow 2$, respectively. $h_a$ has identical model structure with $h_t$.
For all three sub-networks, ReLU activation function and dropout layer with $0.25$ dropout ratio are applied between the two fully connected layers. 
For CelebA dataset, we use ResNet18 as backbone, where the first three stages are used as $f_s$ and the last stage (together with the fully connected classification layer) is used as $h_t$. The auxiliary adversarial head $h_a$ has the same structure as $h_t$.

\paragraph{Baseline methods}
We compare CUMA with the following state-of-the-art in-distribution fairness algorithms. 
Adversarial debiasing (AdvDebias) \citep{zhang2018mitigating} is one of the most popular fair training algorithm based on adversarial training \citep{ganin2016domain}.
\citet{madras2018learning} proposes a similar framework termed Learned Adversarially Fair and Transferable Representations (LAFTR), by replacing the cross-entropy loss used in \citep{zhang2018mitigating} with a group-normalized $\ell_1$ loss, which is shown to work better on highly unbalanced datasets.
We also include normal (fairness-ignorant) training as a baseline. 

\paragraph{Evaluation metric}
We report the \underline{overall accuracy} on all test samples in the original test sets.
To measure \underline{in-distribution fairness}, we use $\Delta_{EO}$ on the original test sets. 
To measure \underline{robust fairness} under distribution shifts, we use $\Delta_{EO}$ on test sets with distribution shifts.
See the following paragraph for the details in constructing distribution shifts. 

\paragraph{Distribution shifts}
Adult dataset contains US Census data collected before 1996. We use the 2014 and 2015 subset of Folktables dataset \cite{ding2021retiring}, which contain US Census data collected in 2014 and 2015 respectively, as the test sets with distribution shifts. 
This simulates the real-world temporal distributional shifts. 

On CelebA dataset, we train the model on $10,000$ ``young'' face images. We use another $1,000$ ``young'' face images as in-distribution test set and $1,000$ ``not young'' face images as the test set with distribution shifts.
This simulates the real-world scenario when the model is trained and used on people with different age groups.
We also construct another test set with a different type of distribution shift, by applying strong JPEG compression on the original $1,000$ ``young'' test images, following \cite{hendrycks2019robustness}. 
This simulates the scenario when the model is trained on good quality images while the test images has poor visual quality.

For C\&C dataset, we construct two artificial distribution shifts by adding random Gaussian and uniform noises, respectively, to the test data. 
Specifically, the categorical features in C\&C dataset are first one-hot encoded and then whitened into float-value vectors, where noises are added. 
Both types of noises have mean $\mu=0$ and has standard derivation $\sigma=0.03$ .

\paragraph{Implementation details}
Unless further specified, we set the loss trade-off parameter $\alpha$ to 1 in all experiments by default. 
We use Adam optimizer \citep{kingma2014adam} with initial learning rate $10^{-3}$ and weight decay $10^{-5}$. The learning rate is gradually decreased to 0 by cosine annealing learning rate scheduler \citep{loshchilov2016sgdr}. 
On both Adult and C\&C datasets, we train for $50$ epochs from scratch for all methods. 
On CelebA dataser, we first normally train a model for 100 epochs, and then finetune it for 20 epochs using CUMA. 
For fair comparison, we train for 120 epochs on CelebA for all baseline methods.
The constant $h$ in Eq.~(\ref{eq:curvature_aprx}) is set to $1$ by default.

\begin{table*}[ht]
\begin{center}
\caption{Results on C\&C dataset with ``RacePctBlack'' as the sensitive attribute. The best and second-best metrics are shown in bold and underlined, respectively. Mean and standard deviation over three random runs are shown for CUMA.}
\vspace{-0.5em}
\resizebox{\linewidth}{!}{
\begin{tabular}{|c|c|c|c|c|}
\hline
\multirow{3}{*}{Method} & \multicolumn{2}{c|}{\makecell{The original C\&C test set}} & \multicolumn{1}{c|}{\makecell{With Gaussian Noise}} & \multicolumn{1}{c|}{\makecell{With Uniform Noise}} \\
\cline{2-5}
& \multirow{2}{*}{Accuracy ($\uparrow$)} &  $\Delta_{EO}$ ($\downarrow$) & $\Delta_{EO}$ ($\downarrow$) & $\Delta_{EO}$ ($\downarrow$) \\
\cline{3-5}
& & \multicolumn{1}{c|}{\textit{In-distribution fairness}} & \multicolumn{2}{c|}{\textit{Robust fairness under distribution shifts}}  \\
\hline\hline
Normal & \textbf{89.05} & 63.22 & 60.13 & 64.21 \\
AdvDebias & {84.79} & 39.84 & 39.84 & 36.81 \\
LAFTR & \underline{85.80} & \underline{28.83} & \underline{29.04} &  \underline{32.20} \\
CUMA & 85.20$\pm1.70$ & \textbf{28.17}$\pm1.70$ &  \textbf{28.69}$\pm1.92$ & \textbf{27.11}$\pm0.82$ \\ 
\hline
\end{tabular}
}
\label{tab:ccrace}
\vspace{-1em}
\end{center}
\end{table*}

\begin{table*}[ht]
\begin{center}
\caption{Results on C\&C dataset with ``FemalePctDiv'' as the sensitive attribute. The best and second-best metrics are shown in bold and underlined, respectively. Mean and standard deviation over three random runs are shown for CUMA.}
\vspace{-0.5em}
\resizebox{\linewidth}{!}{
\begin{tabular}{|c|c|c|c|c|}
\hline
\multirow{3}{*}{Method} & \multicolumn{2}{c|}{\makecell{The original C\&C test set}} & \multicolumn{1}{c|}{\makecell{With Gaussian Noise}} & \multicolumn{1}{c|}{\makecell{With Uniform Noise}} \\
\cline{2-5}
& \multirow{2}{*}{Accuracy ($\uparrow$)} &  $\Delta_{EO}$ ($\downarrow$) & $\Delta_{EO}$ ($\downarrow$) & $\Delta_{EO}$ ($\downarrow$) \\
\cline{3-5}
& & \multicolumn{1}{c|}{\textit{In-distribution fairness}} & \multicolumn{2}{c|}{\textit{Robust fairness under distribution shifts}}  \\
\hline\hline
Normal & \textbf{89.05} & 54.74 & 56.41 &  54.60 \\
AdvDebias & \underline{83.57} & {38.73} & 38.73 & 37.15\\ 
LAFTR & 83.16 & \underline{27.83} &   \underline{29.30} & \underline{30.11}\\
CUMA & 83.39$\pm1.01$ & \textbf{27.57}$\pm0.74$ & \textbf{27.70}$\pm1.04$ & \textbf{28.35}$\pm1.73$ \\ 
\hline
\end{tabular}
}
\label{tab:ccdiv}
\vspace{-1em}
\end{center}
\end{table*}

\subsection{Main Results}
\label{sec:main_results}

Experimental results on three datasets with different sensitive attributes are shown in Tables~\ref{tab:adult}-\ref{tab:ccdiv}, where we compare CUMA with the baseline methods on different metrics as discussed in Section~\ref{sec:settings}.
``Normal'' means standard training without any fairness regularization. 
All numbers are shown as percentages.
Many intriguing findings can be concluded from the results.

\underline{First}, we see that previous state-of-the-art fairness learning algorithms would be jeopardized if distributional shifts are present in test data. 
For example, on Adult dataset (Table~\ref{tab:adult}), LAFTR achieves $\Delta_{EO} = 11.96\%$ on in-distribution test set, while that number is increased to $13.95\%$ on the 2014 test set and $14.80\%$ on the 2015 test set, which is almost as unfair as the normally trained model.
Similarly, on CelenA dataset with ``Chubby" as the sensitive attribute (Table~\ref{tab:celeba_chubby}), LAFTR achieves $\Delta_{EO} = 31.02\%$ on the original CelebA test set, while that number is increased to $35.88\%$  and $37.46\%$ under distribution shifts of user age and image quality, respectively.  

\underline{Second}, we see that CUMA achieves the best robust fairness under distribution shifts under all evaluated settings, while maintaining similar in-distribution fairness and overall accuracy. 
For example, on Adult dataset (Table~\ref{tab:adult}), CUMA achieves $1.92\%$ and $1.96\%$ less $\Delta_{EO}$ than the second-best performer (AdvDebias) on the 2014 and 2015 Census dataset, respectively.
On CelebA dataset (Table~\ref{tab:celeba_chubby}) with ``Chubby" as the sensitive attribute, CUMA achieves $2.62\%$ and $2.74\%$ less $\Delta_{EO}$ than the second-best performers under distribution shifts of user age and image quality, respectively.
Moreover, still in Table~\ref{tab:celeba_chubby}, CUMA and LAFTR achieve almost identical in-distribution fairness (the difference between their $\Delta_{EO}$ on original test set is $0.5\%$), CUMA keeps the fairness under distribution shifts (with only around $1.6\%$ increase in $\Delta_{EO}$), while the fairness achieved by LAFTR is significantly worse, especially under the image quality distribution shift, where the $\Delta_{EO}$ is increased by $6.44\%$. 




\subsection{Ablation Study} \label{sec:abla}

In this section, we check the sensitivity of CUMA with respect to its hyper-parameters: the loss trade-off parameters $\alpha$ and $\gamma$ in Eq.~(\ref{eq:overall_loss}) and $h$ in Eq.~(\ref{eq:curvature_aprx}).
Results are shown in Table~\ref{tab:abla_cuma}.
When fixing $\alpha=1$, the best trade-off between overall accuracy and robust fairness is achieved at round $\gamma=1$, which we use as the default $\gamma$. 
Varying the value of  $h$ hardly affects the performance of CUMA.

\begin{table}[ht]
\centering
\vspace{-1em}
\caption{Ablation study results on the loss trade-off parameters $\alpha$ and $\gamma$ in the CUMA algorithm. Results are reported on C\&C dataset with ``RacePctBlack'' as the sensitive attribute.}
\resizebox{1.0\linewidth}{!}
{
\begin{tabular}{c|ccc|ccc|cc}
\hline
& \multicolumn{3}{c|}{$\alpha$} & \multicolumn{3}{c|}{$\gamma$} & \multicolumn{2}{c}{$h$} \\
 & 0.1 & 1 & 10 & 0.1 & 1 & 10 & 0.1 & 1 \\
\hline
Accuracy & 86.94 & 85.40 & 83.75 & 85.19 & 85.40 & 84.79 &  85.32  & 85.40 \\
\hline
\makecell{$\Delta_{EO}$\\with Gaussian noise} & 66.51 & 28.74 & 33.16 & 38.85 & 28.74 & 27.95 & 30.56 & 28.74 \\
\hline
\end{tabular}
}
\label{tab:abla_cuma}
\end{table}

\subsection{Trade-off Curves between Fairness and Accuracy}
\label{sec:appx-tradeoff-curves}
\begin{figure}[h]
\centering
\setlength{\tabcolsep}{1pt}
\begin{tabular}{ccc}
    \includegraphics[width=0.33\linewidth]{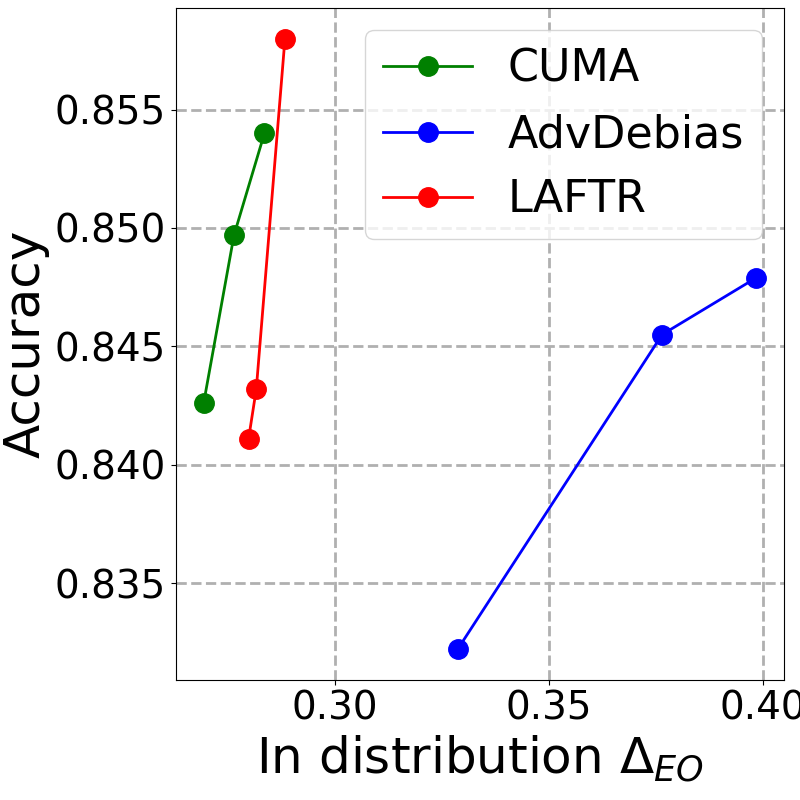} &
	\includegraphics[width=0.33\linewidth]{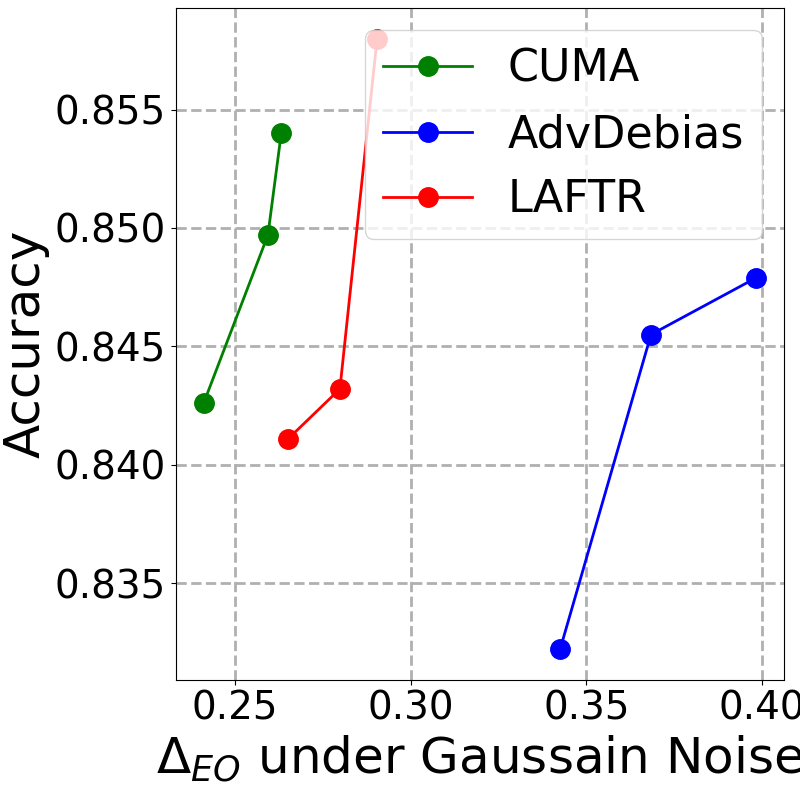} &
	\includegraphics[width=0.33\linewidth]{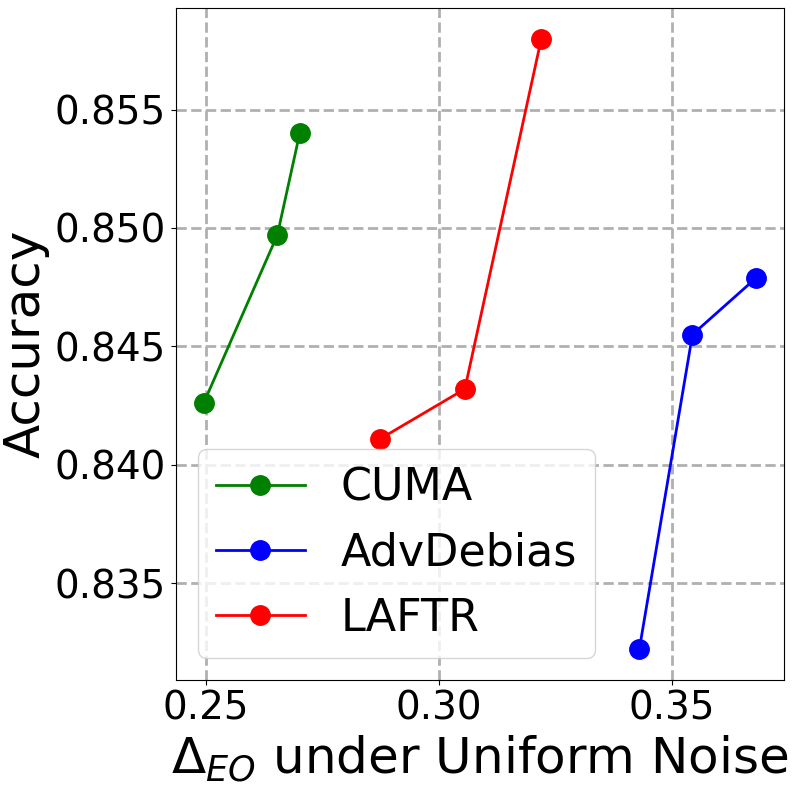}  \\
\end{tabular}
\vspace{-1em}
\caption{Trade-off curves between fairness and accuracy of different methods. Results are reported on C\&C  dataset with ``RacePctBlack'' as the sensitive attribute.}
\label{fig:curves}
\vspace{-1em}
\end{figure}

For CUMA and both baseline methods, we can obtain different trade-offs between fairness and accuracy by setting the loss function weights (e.g., $\alpha$ and $\gamma$) to different values. For example, the larger $\alpha$, the better fairness and the worse accuracy. 
Such trade-off curves between fairness and accuracy of different methods are shown in Figure \ref{fig:curves}.
The closer the curve to the top-left corner (i.e., with larger accuracy and smaller $\Delta_{EO}$), the better Pareto frontier is achieved. 
As we can see, our method achieves the best Pareto frontiers for both in-distribution fairness (left panel) and robust  fairness  under  distribution  shifts (middle and right panel). 

\section{Conclusion}
In this paper, we first observe the challenge of robust fairness: Existing state-of-the-art in-distribution fairness learning methods suffer significant performance drop under unseen distribution shifts. 
To solve this problem, we propose a novel robust fairness learning algorithm, termed Curvature Matching (CUMA), to simultaneously achieve both traditional in-distribution fairness and robust fairness.
Experiments show CUMA achieves more robust fairness under unseen  distribution shifts, without more sacrifice on either overall accuracies or the in-distribution fairness compared with traditional in-distribution fairness learning methods.

\bibliography{reference}

\begin{thebibliography}{60}
\providecommand{\natexlab}[1]{#1}
\providecommand{\url}[1]{\texttt{#1}}
\expandafter\ifx\csname urlstyle\endcsname\relax
  \providecommand{\doi}[1]{doi: #1}\else
  \providecommand{\doi}{doi: \begingroup \urlstyle{rm}\Url}\fi

\bibitem[Agarwal et~al.(2018)Agarwal, Beygelzimer, Dud{\'\i}k, Langford, and
  Wallach]{agarwal2018reductions}
Agarwal, A., Beygelzimer, A., Dud{\'\i}k, M., Langford, J., and Wallach, H.
\newblock A reductions approach to fair classification.
\newblock In \emph{International Conference on Machine Learning}, pp.\  60--69,
  2018.

\bibitem[Bahng et~al.(2020)Bahng, Chun, Yun, Choo, and Oh]{bahng2020learning}
Bahng, H., Chun, S., Yun, S., Choo, J., and Oh, S.~J.
\newblock Learning de-biased representations with biased representations.
\newblock In \emph{International Conference on Machine Learning}, pp.\
  528--539, 2020.

\bibitem[Bartlett et~al.(2017)Bartlett, Foster, and
  Telgarsky]{bartlett2017spectrally}
Bartlett, P., Foster, D.~J., and Telgarsky, M.
\newblock Spectrally-normalized margin bounds for neural networks.
\newblock In \emph{Advances in Neural Information Processing Systems}, 2017.

\bibitem[Bi{\'n}kowski et~al.(2018)Bi{\'n}kowski, Sutherland, Arbel, and
  Gretton]{binkowski2018demystifying}
Bi{\'n}kowski, M., Sutherland, D.~J., Arbel, M., and Gretton, A.
\newblock Demystifying {MMD GANs}.
\newblock In \emph{International Conference on Learning Representations}, 2018.

\bibitem[Buolamwini \& Gebru(2018)Buolamwini and Gebru]{buolamwini2018gender}
Buolamwini, J. and Gebru, T.
\newblock Gender shades: Intersectional accuracy disparities in commercial
  gender classification.
\newblock In \emph{Conference on Fairness, Accountability and Transparency},
  pp.\  77--91, 2018.

\bibitem[Calmon et~al.(2017)Calmon, Wei, Vinzamuri, Ramamurthy, and
  Varshney]{calmon2017optimized}
Calmon, F.~P., Wei, D., Vinzamuri, B., Ramamurthy, K.~N., and Varshney, K.~R.
\newblock Optimized pre-processing for discrimination prevention.
\newblock In \emph{International Conference on Neural Information Processing
  Systems}, pp.\  3995--4004, 2017.

\bibitem[Creager et~al.(2019)Creager, Madras, Jacobsen, Weis, Swersky, Pitassi,
  and Zemel]{creager2019flexibly}
Creager, E., Madras, D., Jacobsen, J.-H., Weis, M., Swersky, K., Pitassi, T.,
  and Zemel, R.
\newblock Flexibly fair representation learning by disentanglement.
\newblock In \emph{International Conference on Machine Learning}, pp.\
  1436--1445, 2019.

\bibitem[Dai \& Brown(2020)Dai and Brown]{dailabel}
Dai, J. and Brown, S.~M.
\newblock Label bias, label shift: Fair machine learning with unreliable
  labels.
\newblock In \emph{NeurIPS Workshop}, 2020.

\bibitem[de~Vries et~al.(2019)de~Vries, Misra, Wang, and van~der
  Maaten]{de2019does}
de~Vries, T., Misra, I., Wang, C., and van~der Maaten, L.
\newblock Does object recognition work for everyone?
\newblock In \emph{IEEE Conference on Computer Vision and Pattern Recognition
  Workshops}, pp.\  52--59, 2019.

\bibitem[Ding et~al.(2021)Ding, Hardt, Miller, and Schmidt]{ding2021retiring}
Ding, F., Hardt, M., Miller, J., and Schmidt, L.
\newblock Retiring {Adult}: New datasets for fair machine learning.
\newblock \emph{arXiv preprint arXiv:2108.04884}, 2021.

\bibitem[Du et~al.(2020)Du, Yang, Zou, and Hu]{du2020fairness}
Du, M., Yang, F., Zou, N., and Hu, X.
\newblock Fairness in deep learning: A computational perspective.
\newblock \emph{IEEE Intelligent Systems}, 2020.

\bibitem[Dwork et~al.(2012)Dwork, Hardt, Pitassi, Reingold, and
  Zemel]{dwork2012fairness}
Dwork, C., Hardt, M., Pitassi, T., Reingold, O., and Zemel, R.
\newblock Fairness through awareness.
\newblock In \emph{Innovations in Theoretical Computer Science Conference},
  pp.\  214--226, 2012.

\bibitem[Edwards \& Storkey(2016)Edwards and Storkey]{edwards2015censoring}
Edwards, H. and Storkey, A.
\newblock Censoring representations with an adversary.
\newblock In \emph{International Conference on Learning Representations}, 2016.

\bibitem[Ganin et~al.(2016)Ganin, Ustinova, Ajakan, Germain, Larochelle,
  Laviolette, Marchand, and Lempitsky]{ganin2016domain}
Ganin, Y., Ustinova, E., Ajakan, H., Germain, P., Larochelle, H., Laviolette,
  F., Marchand, M., and Lempitsky, V.
\newblock Domain-adversarial training of neural networks.
\newblock \emph{Journal of Machine Learning Research}, 17\penalty0
  (1):\penalty0 2096--2030, 2016.

\bibitem[Gretton et~al.(2012)Gretton, Borgwardt, Rasch, Sch{\"o}lkopf, and
  Smola]{gretton2012kernel}
Gretton, A., Borgwardt, K.~M., Rasch, M.~J., Sch{\"o}lkopf, B., and Smola, A.
\newblock A kernel two-sample test.
\newblock \emph{Journal of Machine Learning Research}, 13\penalty0
  (1):\penalty0 723--773, 2012.

\bibitem[Guo et~al.(2018)Guo, Zhang, Zhang, and Chen]{guo2018sparse}
Guo, Y., Zhang, C., Zhang, C., and Chen, Y.
\newblock Sparse {DNNs} with improved adversarial robustness.
\newblock In \emph{Advances in Neural Information Processing Systems}, 2018.

\bibitem[Hardt et~al.(2016)Hardt, Price, and Srebro]{hardt2016equality}
Hardt, M., Price, E., and Srebro, N.
\newblock Equality of opportunity in supervised learning.
\newblock In \emph{Advances in Neural Information Processing Systems}, 2016.

\bibitem[Hashimoto et~al.(2018)Hashimoto, Srivastava, Namkoong, and
  Liang]{hashimoto2018fairness}
Hashimoto, T., Srivastava, M., Namkoong, H., and Liang, P.
\newblock Fairness without demographics in repeated loss minimization.
\newblock In \emph{International Conference on Machine Learning}, pp.\
  1929--1938, 2018.

\bibitem[Hendrycks \& Dietterich(2019)Hendrycks and
  Dietterich]{hendrycks2019robustness}
Hendrycks, D. and Dietterich, T.
\newblock Benchmarking neural network robustness to common corruptions and
  perturbations.
\newblock \emph{International Conference on Learning Representations}, 2019.

\bibitem[Hendrycks et~al.(2020)Hendrycks, Basart, Mu, Kadavath, Wang, Dorundo,
  Desai, Zhu, Parajuli, Guo, Song, Steinhardt, and Gilmer]{hendrycks2020many}
Hendrycks, D., Basart, S., Mu, N., Kadavath, S., Wang, F., Dorundo, E., Desai,
  R., Zhu, T., Parajuli, S., Guo, M., Song, D., Steinhardt, J., and Gilmer, J.
\newblock The many faces of robustness: A critical analysis of
  out-of-distribution generalization.
\newblock \emph{arXiv preprint arXiv:2006.16241}, 2020.

\bibitem[Hendrycks et~al.(2021)Hendrycks, Zhao, Basart, Steinhardt, and
  Song]{hendrycks2021nae}
Hendrycks, D., Zhao, K., Basart, S., Steinhardt, J., and Song, D.
\newblock Natural adversarial examples.
\newblock In \emph{IEEE Conference on Computer Vision and Pattern Recognition},
  2021.

\bibitem[Hwang et~al.(2020)Hwang, Park, Kim, Do, and
  Byun]{hwang2020fairfacegan}
Hwang, S., Park, S., Kim, D., Do, M., and Byun, H.
\newblock Fairfacegan: Fairness-aware facial image-to-image translation.
\newblock In \emph{British Machine Vision Conference}, 2020.

\bibitem[Kamiran \& Calders(2012)Kamiran and Calders]{kamiran2012data}
Kamiran, F. and Calders, T.
\newblock Data preprocessing techniques for classification without
  discrimination.
\newblock \emph{Knowledge and Information Systems}, 33\penalty0 (1):\penalty0
  1--33, 2012.

\bibitem[Kingma \& Ba(2014)Kingma and Ba]{kingma2014adam}
Kingma, D.~P. and Ba, J.
\newblock Adam: A method for stochastic optimization.
\newblock \emph{arXiv preprint arXiv:1412.6980}, 2014.

\bibitem[Kingma \& Welling(2013)Kingma and Welling]{kingma2013auto}
Kingma, D.~P. and Welling, M.
\newblock Auto-encoding variational bayes.
\newblock \emph{arXiv preprint arXiv:1312.6114}, 2013.

\bibitem[Kohavi(1996)]{kohavi1996scaling}
Kohavi, R.
\newblock Scaling up the accuracy of {naive-Bayes} classifiers: A decision-tree
  hybrid.
\newblock In \emph{International Conference on Knowledge Discovery and Data
  Mining}, pp.\  202--207, 1996.

\bibitem[Li et~al.(2017)Li, Chang, Cheng, Yang, and P{\'o}czos]{li2017mmd}
Li, C.-L., Chang, W.-C., Cheng, Y., Yang, Y., and P{\'o}czos, B.
\newblock {MMD GAN}: Towards deeper understanding of moment matching network.
\newblock In \emph{International Conference on Machine Learning}, 2017.

\bibitem[Li et~al.(2019)Li, Sanjabi, Beirami, and Smith]{li2019fair}
Li, T., Sanjabi, M., Beirami, A., and Smith, V.
\newblock Fair resource allocation in federated learning.
\newblock In \emph{International Conference on Learning Representations}, 2019.

\bibitem[Li \& Vasconcelos(2019)Li and Vasconcelos]{li2019repair}
Li, Y. and Vasconcelos, N.
\newblock {REPAIR}: Removing representation bias by dataset resampling.
\newblock In \emph{IEEE Conference on Computer Vision and Pattern Recognition},
  pp.\  9572--9581, 2019.

\bibitem[Li et~al.(2015)Li, Swersky, and Zemel]{li15gmmn}
Li, Y., Swersky, K., and Zemel, R.
\newblock Generative moment matching networks.
\newblock In \emph{International Conference on Machine Learning}, pp.\
  1718--1727, 2015.

\bibitem[Liu et~al.(2015)Liu, Luo, Wang, and Tang]{liu2015faceattributes}
Liu, Z., Luo, P., Wang, X., and Tang, X.
\newblock Deep learning face attributes in the wild.
\newblock In \emph{IEEE International Conference on Computer Vision}, 2015.

\bibitem[Loshchilov \& Hutter(2016)Loshchilov and Hutter]{loshchilov2016sgdr}
Loshchilov, I. and Hutter, F.
\newblock {SGDR}: Stochastic gradient descent with warm restarts.
\newblock \emph{arXiv preprint arXiv:1608.03983}, 2016.

\bibitem[Madras et~al.(2018)Madras, Creager, Pitassi, and
  Zemel]{madras2018learning}
Madras, D., Creager, E., Pitassi, T., and Zemel, R.
\newblock Learning adversarially fair and transferable representations.
\newblock In \emph{International Conference on Machine Learning}, pp.\
  3384--3393, 2018.

\bibitem[Mandal et~al.(2020)Mandal, Deng, Jana, and Hsu]{mandal2020ensuring}
Mandal, D., Deng, S., Jana, S., and Hsu, D.
\newblock Ensuring fairness beyond the training data.
\newblock In \emph{NeurIPS}, 2020.

\bibitem[Martinez et~al.(2020)Martinez, Bertran, and
  Sapiro]{martinez2020minimax}
Martinez, N., Bertran, M., and Sapiro, G.
\newblock Minimax {Pareto} fairness: A multi objective perspective.
\newblock In \emph{International Conference on Machine Learning}, pp.\
  6755--6764, 2020.

\bibitem[Miyato et~al.(2018)Miyato, Maeda, Koyama, and
  Ishii]{miyato2018virtual}
Miyato, T., Maeda, S.-i., Koyama, M., and Ishii, S.
\newblock Virtual adversarial training: a regularization method for supervised
  and semi-supervised learning.
\newblock \emph{IEEE Transactions on Pattern Analysis and Machine
  Intelligence}, 41\penalty0 (8):\penalty0 1979--1993, 2018.

\bibitem[Moosavi-Dezfooli et~al.(2019)Moosavi-Dezfooli, Fawzi, Uesato, and
  Frossard]{moosavi2019robustness}
Moosavi-Dezfooli, S.-M., Fawzi, A., Uesato, J., and Frossard, P.
\newblock Robustness via curvature regularization, and vice versa.
\newblock In \emph{IEEE Conference on Computer Vision and Pattern Recognition},
  pp.\  9078--9086, 2019.

\bibitem[Mu\~{n}oz et~al.(2016)Mu\~{n}oz, Smith, and Patil]{executive2016big}
Mu\~{n}oz, C., Smith, M., and Patil, D.
\newblock \emph{Big data: A report on algorithmic systems, opportunity, and
  civil rights}.
\newblock United States Executive Office of the President, 2016.

\bibitem[Nagpal et~al.(2019)Nagpal, Singh, Singh, and Vatsa]{nagpal2019deep}
Nagpal, S., Singh, M., Singh, R., and Vatsa, M.
\newblock Deep learning for face recognition: Pride or prejudiced?
\newblock \emph{arXiv preprint arXiv:1904.01219}, 2019.

\bibitem[Nam et~al.(2020)Nam, Cha, Ahn, Lee, and Shin]{nam2020learning}
Nam, J., Cha, H., Ahn, S.-S., Lee, J., and Shin, J.
\newblock Learning from failure: De-biasing classifier from biased classifier.
\newblock In \emph{Advances in Neural Information Processing Systems}, 2020.

\bibitem[Park et~al.(2020)Park, Hwang, Hong, and Byun]{park2020fair}
Park, S., Hwang, S., Hong, J., and Byun, H.
\newblock {Fair-VQA}: Fairness-aware visual question answering through
  sensitive attribute prediction.
\newblock \emph{IEEE Access}, 8:\penalty0 215091--215099, 2020.

\bibitem[Podesta et~al.(2014)Podesta, Pritzker, Moniz, Holdren, and
  Zients]{housebig}
Podesta, J., Pritzker, P., Moniz, E.~J., Holdren, J., and Zients, J.
\newblock \emph{Big data: Seizing opportunities and preserving values}.
\newblock United States Executive Office of the President, 2014.

\bibitem[Qiu et~al.(2019)Qiu, Leng, Guo, Chen, Li, Guo, and
  Zhu]{qiu2019adversarial}
Qiu, Y., Leng, J., Guo, C., Chen, Q., Li, C., Guo, M., and Zhu, Y.
\newblock Adversarial defense through network profiling based path extraction.
\newblock In \emph{IEEE Conference on Computer Vision and Pattern Recognition},
  pp.\  4777--4786, 2019.

\bibitem[Quadrianto \& Sharmanska(2017)Quadrianto and
  Sharmanska]{quadrianto2017recycling}
Quadrianto, N. and Sharmanska, V.
\newblock Recycling privileged learning and distribution matching for fairness.
\newblock In \emph{Advances in Neural Information Processing Systems}, 2017.

\bibitem[Quadrianto et~al.(2019)Quadrianto, Sharmanska, and
  Thomas]{quadrianto2019discovering}
Quadrianto, N., Sharmanska, V., and Thomas, O.
\newblock Discovering fair representations in the data domain.
\newblock In \emph{IEEE Conference on Computer Vision and Pattern Recognition},
  pp.\  8227--8236, 2019.

\bibitem[Redmond \& Baveja(2002)Redmond and Baveja]{redmond2002data}
Redmond, M. and Baveja, A.
\newblock A data-driven software tool for enabling cooperative information
  sharing among police departments.
\newblock \emph{European Journal of Operational Research}, 141\penalty0
  (3):\penalty0 660--678, 2002.

\bibitem[Rezaei et~al.(2021)Rezaei, Liu, Memarrast, and
  Ziebart]{rezaei2021robust}
Rezaei, A., Liu, A., Memarrast, O., and Ziebart, B.~D.
\newblock Robust fairness under covariate shift.
\newblock In \emph{AAAI}, pp.\  9419--9427, 2021.

\bibitem[Sarhan et~al.(2020)Sarhan, Navab, Eslami, and
  Albarqouni]{sarhan2020fairness}
Sarhan, M.~H., Navab, N., Eslami, A., and Albarqouni, S.
\newblock Fairness by learning orthogonal disentangled representations.
\newblock In \emph{European Conference on Computer Vision}, pp.\  746--761,
  2020.

\bibitem[Shankar et~al.(2017)Shankar, Halpern, Breck, Atwood, Wilson, and
  Sculley]{shankar2017no}
Shankar, S., Halpern, Y., Breck, E., Atwood, J., Wilson, J., and Sculley, D.
\newblock No classification without representation: Assessing geodiversity
  issues in open data sets for the developing world.
\newblock In \emph{Advances in Neural Information Processing Systems Workshop},
  2017.

\bibitem[Singh et~al.(2021)Singh, Singh, Mhasawade, and
  Chunara]{singh2021fairness}
Singh, H., Singh, R., Mhasawade, V., and Chunara, R.
\newblock Fairness violations and mitigation under covariate shift.
\newblock In \emph{FAccT}, pp.\  3--13, 2021.

\bibitem[Smuha(2019)]{smuha2019eu}
Smuha, N.~A.
\newblock The {EU} approach to ethics guidelines for trustworthy artificial
  intelligence.
\newblock \emph{Computer Law Review International}, 20\penalty0 (4):\penalty0
  97--106, 2019.

\bibitem[Taori et~al.(2020)Taori, Dave, Shankar, Carlini, Recht, and
  Schmidt]{taori2020measuring}
Taori, R., Dave, A., Shankar, V., Carlini, N., Recht, B., and Schmidt, L.
\newblock Measuring robustness to natural distribution shifts in image
  classification.
\newblock In \emph{Advances in Neural Information Processing Systems}, 2020.

\bibitem[Wadsworth et~al.(2018)Wadsworth, Vera, and
  Piech]{wadsworth2018achieving}
Wadsworth, C., Vera, F., and Piech, C.
\newblock Achieving fairness through adversarial learning: an application to
  recidivism prediction.
\newblock \emph{arXiv preprint arXiv:1807.00199}, 2018.

\bibitem[Wang et~al.(2019)Wang, Zhao, Yatskar, Chang, and
  Ordonez]{wang2019balanced}
Wang, T., Zhao, J., Yatskar, M., Chang, K.-W., and Ordonez, V.
\newblock Balanced datasets are not enough: Estimating and mitigating gender
  bias in deep image representations.
\newblock In \emph{IEEE International Conference on Computer Vision}, pp.\
  5310--5319, 2019.

\bibitem[Weng et~al.(2018)Weng, Zhang, Chen, Yi, Su, Gao, Hsieh, and
  Daniel]{weng2018evaluating}
Weng, T.-W., Zhang, H., Chen, P.-Y., Yi, J., Su, D., Gao, Y., Hsieh, C.-J., and
  Daniel, L.
\newblock Evaluating the robustness of neural networks: An extreme value theory
  approach.
\newblock In \emph{International Conference on Learning Representations}, 2018.

\bibitem[Wilson et~al.(2019)Wilson, Hoffman, and
  Morgenstern]{wilson2019predictive}
Wilson, B., Hoffman, J., and Morgenstern, J.
\newblock Predictive inequity in object detection.
\newblock \emph{arXiv preprint arXiv:1902.11097}, 2019.

\bibitem[Yan et~al.(2021)Yan, Zhang, Niu, Feng, Tan, and Sugiyama]{yan2021cifs}
Yan, H., Zhang, J., Niu, G., Feng, J., Tan, V.~Y., and Sugiyama, M.
\newblock {CIFS}: Improving adversarial robustness of cnns via channel-wise
  importance-based feature selection.
\newblock \emph{arXiv preprint arXiv:2102.05311}, 2021.

\bibitem[Zhang et~al.(2018)Zhang, Lemoine, and Mitchell]{zhang2018mitigating}
Zhang, B.~H., Lemoine, B., and Mitchell, M.
\newblock Mitigating unwanted biases with adversarial learning.
\newblock In \emph{AAAI Conference on AI, Ethics, and Society}, pp.\  335--340,
  2018.

\bibitem[Zhang et~al.(2021)Zhang, Bifet, Zhang, Weiss, and
  Nejdl]{zhang2021farf}
Zhang, W., Bifet, A., Zhang, X., Weiss, J.~C., and Nejdl, W.
\newblock {FARF}: A fair and adaptive random forests classifier.
\newblock In \emph{PACKDD}, pp.\  245--256, 2021.

\bibitem[Zhao et~al.(2017)Zhao, Wang, Yatskar, Ordonez, and Chang]{zhao2017men}
Zhao, J., Wang, T., Yatskar, M., Ordonez, V., and Chang, K.-W.
\newblock Men also like shopping: Reducing gender bias amplification using
  corpus-level constraints.
\newblock \emph{arXiv preprint arXiv:1707.09457}, 2017.

\end{thebibliography}
\bibliographystyle{icml2022}

\end{document}